\documentclass[preprint,12pt]{elsarticle}

\usepackage[title]{appendix}
\usepackage{amssymb}
\usepackage{times}
\usepackage{epsfig}
\usepackage{graphicx}
\usepackage{amsmath}
\usepackage{amssymb}
\usepackage{booktabs}
\usepackage{color}
\usepackage{threeparttable} 
\usepackage{multirow}
\usepackage{bbding}
\DeclareMathOperator*{\argmax}{arg\,max}
\DeclareMathOperator*{\argmin}{arg\,min}
\DeclareMathOperator{\tr}{tr}

\usepackage{url} 
\usepackage{hyperref}

\journal{Nuclear Physics B}

\begin{document}

\begin{frontmatter}

\title{Coping with Change: Learning Invariant and Minimum Sufficient Representations for Fine-Grained Visual Categorization}

\author[inst1]{Shuo Ye}
\affiliation[inst1]{organization={School of Electronic Information and Communications},%Department and Organization
            addressline={Huazhong University of Science and Technology}, 
            city={Wuhan},
            postcode={430074}, 
            state={Huibei},
            country={China}}

\author[inst2,inst3]{Shujian Yu \corref{cor1}}
\author[inst1]{Wenjin Hou}
\author[inst1]{Yu Wang}
\author[inst1]{Xinge You \corref{cor1}}

\affiliation[inst2]{organization={Department of Computer Science},%Department and Organization
            addressline={Vrije Universiteit Amsterdam}, 
            city={Amsterdam},
            country={The Netherlands}}

\affiliation[inst3]{organization={Machine Learning Group},%Department and Organization
            addressline={UiT - The Arctic University of Norway}, 
            city={Troms{\o}},
            country={Norway}}
                    
\cortext[cor1]{Corresponding author}

\begin{abstract}
Fine-grained visual categorization~(FGVC) is a challenging task due to similar visual appearances between various species. 
Previous studies always implicitly assume that the training and test data have the same underlying distributions, and that features extracted by modern backbone architectures remain discriminative and generalize well to unseen test data.
However, we empirically justify that these conditions are not always true on benchmark datasets.
To this end, we combine the merits of invariant risk minimization (IRM) and information bottleneck (IB) principle to learn invariant and minimum sufficient~(IMS) representations for FGVC, such that the overall model can always discover the most succinct and consistent fine-grained features. 
We apply the matrix-based R{\'e}nyi's $\alpha$-order entropy to simplify and stabilize the training of IB; we also design a ``soft" environment partition scheme to make IRM applicable to FGVC task. To the best of our knowledge, we are the first to address the problem of FGVC from a generalization perspective and develop a new information-theoretic solution accordingly. 
Extensive experiments demonstrate the consistent performance gain offered by our IMS.
Code is available at: https://github.com/SYe-hub/IMS
\end{abstract}




\end{frontmatter}


\section{Introduction}
Fine-grained visual categorization (FGVC) aims to accurately identify the subordinate classes~(e.g., \textit{Afterglow} and \textit{Rubromarginata}) from a target class~(e.g., flowers),  providing people with more precise and intelligent visual perception capabilities. 
It has a wide range of applications in various industrial and practical domains, such as public safety (e.g., distinguishing between two highly similar vehicles in a complex traffic environment, such as the \textit{acura RL} and \textit{acura TL}.) and retail~(e.g., packaging similar products with vastly different prices, such as \textit{Alcohol} and \textit{Seasoner}.)~\cite{ye2023image}. 
Compared to general image classification, fine-grained images exhibit highly similar appearances and are vulnerable to factors such as pose and viewpoint variations. 
This leads to a small inter-class variance and a large intra-class variance, posing challenges in accurately distinguishing targets.
%Unlike general image classification tasks where deep models are used to automatically represent images, fine-grained images exhibit highly similar appearances and are susceptible to factors such as pose and viewpoint. Moreover, during the data collection process, factors like lighting and occlusion interfere and often lead to incomplete expression of discriminative local features, making the automatic learning of features particularly challenging.
Therefore, the key to addressing this problem is to identify fine-grained visual differences that are sufficient to discriminate similar objects~\cite{wei2019rpc,liu2016deepfashion,chen2020fine}.

Earlier approaches~\cite{zhang2014part,zhou2016learning} make use of bounding box and part annotations to manually identify fine-grained regions or features. However, hand labeling is a labor-intensive and tedious job. Moreover, the discriminant area of human subjective annotation is not always suitable for the computer~\cite{wei2021fine,ye2022discriminative}. 
Recent efforts~\cite{liu2022transformer,ye2022discriminative} focus on recognition with only image-level labels. 
These methods typically apply a pre-trained network as a backbone and demonstrate noticeable performance improvements.

\begin{figure}[t]
  \begin{center}
  \includegraphics[width=0.87\textwidth]{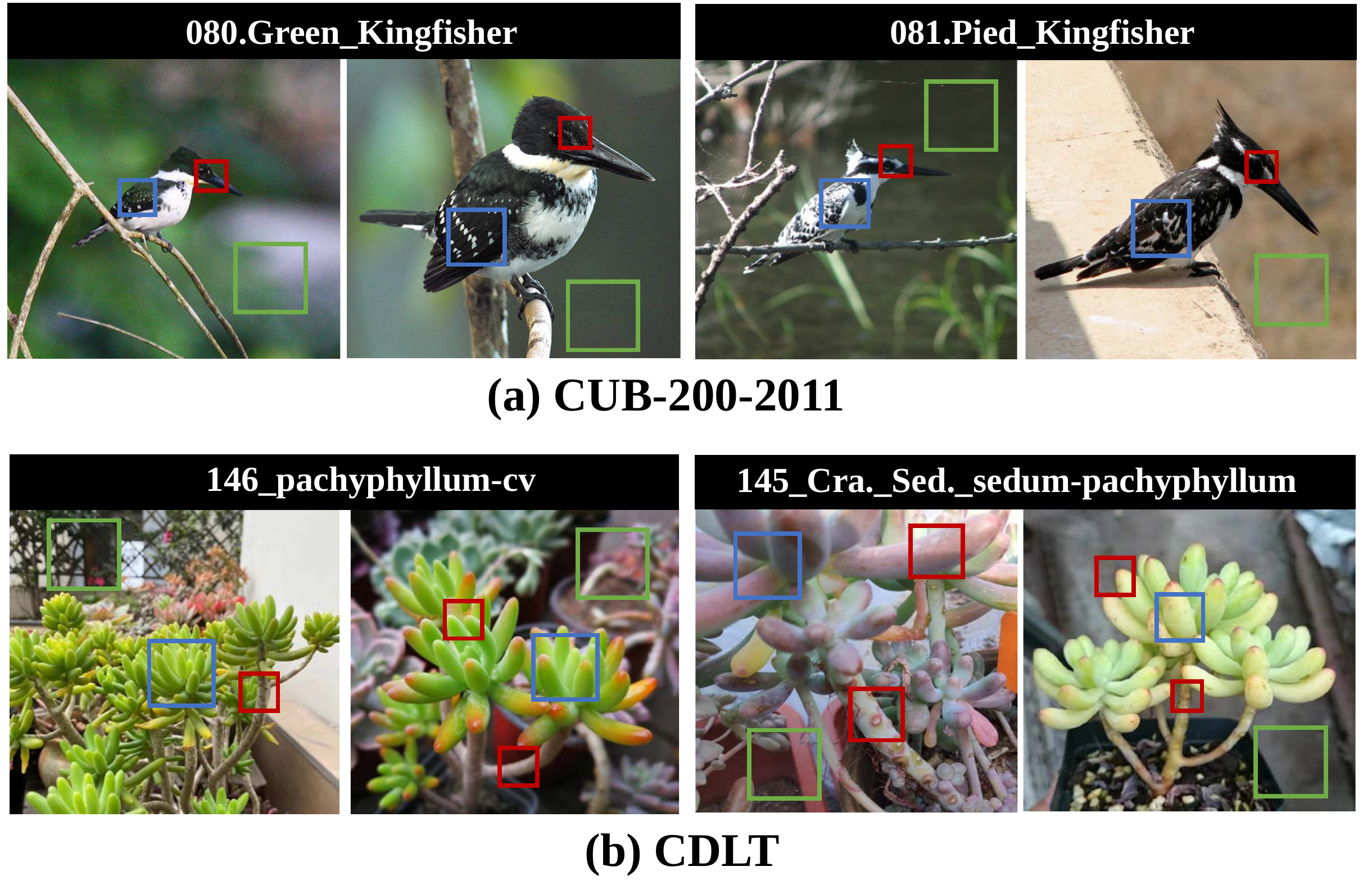}
  \end{center}
	\setlength{\abovecaptionskip}{-0.2cm} 
  \vspace{-0.4cm}
  \caption{Distributional shift in FGVC can lead models to rely on spurious correlations~(e.g., color) and irrelevant information~(e.g., background) when making decisions.
Invariant, spurious, and redundant features are marked using {\color{red} red}, {\color{blue} blue}, and {\color{green} green} boxes, respectively. 
Our model aims to identify only the invariant and minimal sufficient~(IMS) representations for classification.}
  \label{motivation}
   \vspace{-0.2cm}
\end{figure}

Unfortunately, data in reality typically changes over time, which presents a new challenge for computer vision in learning under distributional shift. 
Existing deep learning-based methods are incapable of dealing with such distributional shift because they implicitly assume that the dependence between the extracted features and the class labels remains stable~\cite{zhang2021deep}.
This may result in the decision process taking into account the influence of external factors, such as instance color and environmental illumination.
This problem is especially severe for FGVC, where part of the features may be well-correlated with labels in the training data but become degraded or changed in the test data.
Besides, the extracted representations often contain a large amount of redundant or task-irrelevant information, which increases the risk of ``real" discriminative features being suppressed or underutilized~\cite{wang2022r2,he2022transfg,hu2021rams}.

Fig.~\ref{motivation} illustrates this problem on the benchmark CUB dataset~\cite{wah2011caltech} and CDLT dataset. 
Taking Fig.~\ref{motivation}(a) as an example, 
the first two images and the third image belong to different subclasses, which can be easily distinguished by the color or the texture patterns of the wings~(marked with blue boxes). 
However, these features are not always reliable, as can be seen in the last image where the color and texture pattern of \textit{pied kingfisher} is more similar to that of \textit{green kingfisher}.
In order to remain consistent discrimination power, practitioners are expected to make decisions based on the bird's eye area~(marked with red boxes)\footnote{https://species.sciencereading.cn/biology/v/biologicalIndex/122.html}.
Similarly, in Fig.~\ref{motivation}(b).
the color of leaves cannot always provide a stable discriminant clue, and experienced experts would like to use the structural features of the leaf tips and stems~(marked with red boxes) to distinguish \textit{pachyphyllum.cv} from \textit{pachyphyllum}.
Unfortunately, deep learning models struggle with this, because they are unable to understand the difference between spurious correlations and invariant features when fitting~\cite{arjovsky2019invariant,ahuja2020invariant}.
Furthermore, images often contain redundant or task-irrelevant information~(marked with green boxes), such as visually similar but meaningless backgrounds. 
In summary, the distributional shift of test data and the existence spurious correlation and redundancy in extracted features pose new challenges for FGVC.

\begin{figure}[htbp]
  \begin{center}
  \includegraphics[width=0.87\textwidth]{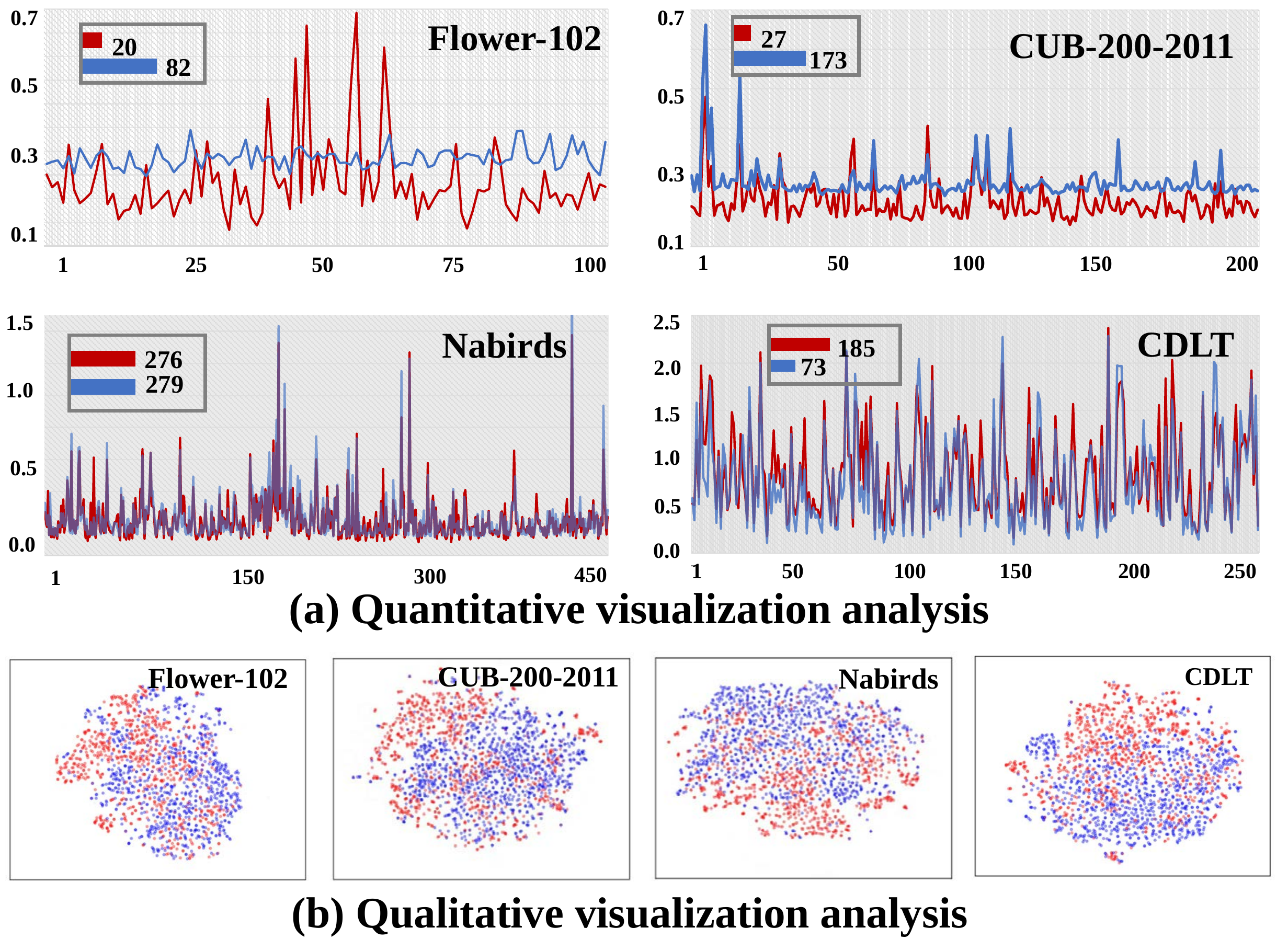}
  \end{center}
	\setlength{\abovecaptionskip}{-0.2cm} 
 \vspace{-0.2cm}
  \caption{Visualization analysis. 
In (a), the red line represents the MMD between the training and test set, and the blue line represents the confidence values that accept the hypothesis.
In (b), t-SNE~\cite{van2008visualizing} is used, where red samples correspond to training set instances, whereas blue samples correspond to test set instances.}
\vspace{-0.2cm}
  \label{Shift_quantization}
\end{figure}

To further justify the phenomena of distribution shift in FGVC, we employ maximum mean discrepancy~(MMD)~\cite{gretton2012kernel} to quantitatively assess the distributional shift in four benchmark FGVC datasets~\cite{wah2011caltech, van2015building, nilsback2008automated,ye2023cdlt}. We use a pre-trained ResNet-50 for feature extraction.
The results are shown in Fig.~\ref{Shift_quantization}, in which we compute the MMD values for each subclass between training and test sets, as shown in the red curve. 
The upper 95\% confidence interval bound (evaluated with 100 independent permutations, see details in~\cite{gretton2012kernel}) is shown as the blue line.
We count the number of subclasses that the MMD value is above the upper 95\% confidence interval, which can be interpreted as having a ``significant" distributional change.
It can be found that distributional shift is a universal problem, especially for Nabirds and CDLT.
We also provide t-SNE~\cite{van2008visualizing} visualizations to further qualitatively demonstrate the distributional shift in the FGVC datasets. To improve the quality of the visualization, we have chosen to display a limited number of subclasses from each dataset, as the number of subclasses varies across the datasets. Specifically, for the four datasets~(Flower-102, CUB, Nabirds, and CDLT), we have selected 40, 30, 20, and 20 subclasses, respectively. (More analysis on data distribution can be found in Appendix A.)

To learn invariant features that maintain stable discriminability and provide reliable clues in different scenes or over different time periods~\cite{li2022invariant}, we propose a novel approach for FGVC to only learn invariant and minimum sufficient (IMS) features. Our IMS mitigates the issues mentioned above and can be generalized to popular backbone architectures.
Specifically, motivated by the recent progress in domain generalization~\cite{zhou2022domain}, the invariant risk minimization~(IRM)~\cite{arjovsky2019invariant} and information bottleneck~(IB)~\cite{tishby2000information} principle are combined together to encourage models to learn the most succinct features that are consistently discriminative across unknown distributional shifts.
However, directly applying IRM to FGVC is not straightforward, as it requires predetermined environment partitions, whereas FGVC data only contains a single training set.
Furthermore, although IB theory has been studied and used for redundant information reduction, explicitly implementing this idea is still challenging for image data, due to the difficulty of estimating mutual information or conditional mutual information terms in high-dimensional spaces.
We address these challenges and demonstrate how to make both IRM and IB applicable to FGVC.
To the best of our knowledge, we are the first to address the problem of FGVC from a generalization perspective. A novel information-theoretic approach is developed accordingly. Our main contributions are threefold:

\begin{enumerate} 
\item We analyze the limitations of FGVC from a generalization perspective, and identify the necessity of learning ``invariant" and ``minimum sufficient" features. 
\item A straightforward application of IRM and IB is infeasible since IRM requires predetermined environment partitions and the difficulty of estimating mutual information in high-dimensional space that is also suitable for minibatch-based optimization.
To this end, we apply the recently proposed matrix-based R{\'e}nyi's $\alpha$-order entropy functional~\cite{giraldo2014measures} to simplify and stabilize the training of IB; we also design a ``soft" environment partition scheme to make IRM applicable to FGVC.
\item Experimental results suggest that our IMS outperforms other state-of-the-art (SOTA) approaches on seven benchmark datasets, regardless of the used backbones.
\end{enumerate}

The rest of this paper is organized as follows. In Sec.~2, we introduce related work, including the popular approaches for FGVC, the basic theory of domain generalization and IRM, as well as related methods of invariant feature learning. Our method will be described in Sec.~3. Experimental results will be presented in Sec.~4. Finally, Sec.~5 will draw a conclusion.

\section{Background Knowledge}
\subsection{Popular FGVC Approaches}
Current weakly supervised FGVC methods can be divided into two categories: multi-stage learning~\cite{liu2022transformer,wang2021feature} and end-to-end learning~\cite{zhu2017b,yu2018hierarchical,chang2020devil}.
Based on the pre-trained backbone network, the former captures possible discriminative regions of the target in the input space~(such as the wings and claws of birds) and then magnifies these regions for learning.
Notable examples in this category include RA-CNN~\cite{fu2017look} and MGE~\cite{zhang2019learning}.
The latter achieves fine-grained learning of the target by specifically learning the high-response regions in the image.
Typical methods include HBP~\cite{yu2018hierarchical} and DSE~\cite{ye2022discriminative}. 
Recently, the vision transformer (ViT)~\cite{dosovitskiy2020image,liu2022transformer,hu2021rams,miao2021complemental} has attempted to leverage the power of self-attention mechanisms to solve the FGVC problem.
However, using ViT directly in FGVC may lead to feature redundancy and high dependence. This is mainly because in the training phase, all tokens have interacted, but not every token has a target associated with it.
This viewpoint has also been implicitly expressed in recent ViT-based papers~\cite{he2022transfg,hu2021rams}.
More importantly, existing methods can only explore the discriminative features of instances, but cannot ensure that the learned features are invariant. 
Unfortunately, some features tend to change over time, and since FGVC exploits subtle differences between instances, such shifts can lead to suboptimal performance.
This problem is defined as distributional shift~\cite{wei2021fine}.

\subsection{Domain Generalization for FGVC}
Domain generalization~(DG) aims to achieve out-of-distribution~(OOD) generalization by utilizing only the source data for model learning~\cite{zhou2022domain}. This task poses challenges for deep learning methods since most learning algorithms heavily rely on the assumption of i.i.d. data from the source domain~\cite{lv2022causality}.
%{\color{blue}Compared to Domain adaptation~(DA), which is also aimed at addressing distributional shift between source and target domains, DG presents greater challenges. One of the reasons for this heightened challenge lies in the fact that DG lacks visibility into target domain data during the training phase.}
In the problem of DG, we have a set of $K\geq 2$ related labeled source domains $\{\mathcal D_{i}\}$$_{i=1}^{K}$. 
A domain (which is also called environment) $e$ with $n_{e}$ training samples drawing i.i.d. from distribution $\mathcal{P}_e$ is denoted by 
$\mathcal{D}_{e}:=\{(\mathbf{x}^{e}_{i},y^{e}_{i})\}^{n_{e}}_{i=1}$, in which $e\in \mathcal{E}_{tr}$ and $\mathcal{E}_{tr}$ refers to the set of training environments. 
Our goal is to train a predictor $y = f(\mathbf{x})$ from the $K$ source domain datasets $\{\mathcal D_{i}\}$$_{i=1}^{K}$ such that it can perform well to an unseen, but related test environment $e\in \mathcal{E}_{all}\backslash \mathcal{E}_{tr}$, in which $\mathcal{E}_{all}$ is the set of all possible environments.
We wish to minimize the maximum risk over all domains $\mathcal{E}_{all}$:
\begin{equation}
\min \max R^{e}(f), e\in \mathcal{E}_{all},
\end{equation}
where $R^{e}:=\mathbb{E}_{(\mathbf{x},y)\sim\mathcal{P}_e }[\ell(f(\mathbf{x}),y)]$ is the risk under environment $e$, $\ell$ denotes a differentiable loss function.

In general, learning a robust predictor that is invariant across different domains with different distributions is challenging~\cite{li2022invariant}. 
%The earliest studies on fine-grained domain adaptation employed attributes~\cite{gebru2017fine} or detailed annotations (e.g., bounding boxes and part landmarks) ~\cite{xu2016webly} to achieve alignment between the source and target domains. Cui et al.~\cite {cui2018large} introduced a method for estimating domain similarity, selecting a subset with high similarity to the target domain from the source domain (large-scale datasets, e.g., iNaturalist2017~\cite{van2018inaturalist}), and utilizing the subset data for data augmentation. Han et al.~\cite{han2022bin} further improved upon the work of~\cite{cui2018large}, enhancing the accuracy of final classification by eliminating interfering images from large-scale datasets. While the previously mentioned methods have effectively addressed the challenges arising from distributional shifts between the training and testing sets, they all necessitate acquiring supplementary supervisory information and a portion of data from the target domain. 
Perhaps one of the most popular approaches is domain augmentation, which aims to generate more training data from simulated environments to improve generalization in real environments. This involves augmenting or generating data with different colors, shapes, positions, textures, etc. ~\cite{yue2019domain,prakash2019structured,yun2019cutmix,zhang2017mixup}. However, data augmentation is not suitable for FGVC (see more details in Section 4.5). This is because FGVC classifier solely depends on local and subtle differences of instances in different categories, whereas existing manipulation techniques are hard to concentrate on these areas.
Recently, PAN~\cite{wang2020progressive} achieves DG by hierarchically constructing labels for source domain instances, aligning the distribution between subclasses at the levels of species, genus, family, and order. In this process, self-attention is designed to capture crucial discriminative regions~\cite{wang2020adversarial}. LADS~\cite{dunlap2022using} attempted to introduce the text modality, leveraging the knowledge of pre-trained large models (e.g., CLIP~\cite{radford2021learning}) to guide the learning of the transformation from the source to the target domain for images.
In this paper, we choose domain invariant representation learning approach~\cite{greenfeld2020robust, yu2021measuring, lu2021metadata, jin2021style, arjovsky2019invariant, liao2022decorr}, which aims to learn representations with either invariant (marginal) distributions  $p(\Phi(x_{i}))$ (e.g., \cite{ganin2015unsupervised, ganin2016domain, long2018conditional}) or feature-conditioned label distributions $p(y_{i}|\Phi(x_{i}))$ (e.g., \cite{li2018deep,tachet2020domain}).
It should be noted that our method attains the capacity to acquire invariant features, maintaining stable discriminability and offering reliable cues across diverse scenes or time periods, all without dependence on any form of source domain information augmentation (e.g., hierarchical or modality information).

\subsection{Invariant Risk Minimization}
The recently developed IRM seeks the invariance of $p(y_{i}|\Phi(x_{i}))$. Specifically, it expects to find an invariant causal predictor $f= g\circ \Phi$ by the following objective:
\begin{equation}\label{1}\small
\text {min}_{g, \Phi} \sum_{e} R^{e}(g \circ \Phi),
\end{equation}
\begin{equation}\label{2}\small
\text{s.t.}, g\in \argmin_{\hat{g}} R^{e}(\hat{g} \circ \Phi),
\end{equation}
where $\Phi$ is a feature extractor that maps input $\mathbf{x}$ into latent representation $\Phi (\mathbf{x})$, and $g$ is a linear classifier. Arjovsky \emph{et al}.~\cite{arjovsky2019invariant} instantiate IRM into the practical version IRMv1:
\begin{equation}\label{eqIRM}\small
\text {min}_{\Phi} \sum_{e} R^{e}(\Phi) + \eta ||\nabla_{g|g=1} R^{e}(g\circ \Phi)||_{2}^{2},
\end{equation}
where $g$=1.0 is a scalar and fixed dummy classifier. 

IRM has gained popularity in recent years and
inspired a line of excellent works, such as IRMG~\cite{ahuja2020invariant} and nonlinear IRM~\cite{lu2021nonlinear}. Theoretically, \cite{rosenfeld2020risks} shows that the performances of IRM can only be guaranteed if a sufficient number of environments are given, and the authors also argue that IRM may become defective. Hence, Ahuja \emph{et al}.~\cite{ahuja2021invariance} introduce additional constraints by combining IRM with IB principle and propose IB-IRM.
Let $I(\cdot ; \cdot)$ denote the mutual information,
IB requires the network to compress the information that representation $\Phi(\mathbf{x})$ contains about input $\mathbf{x}$, while ensuring its predictive power to label $y$ by $\max {I(y; \Phi(\mathbf{x}))-\lambda I(\mathbf{x}, \Phi(\mathbf{x}))}$, in which $\lambda$ is a Lagrange multiplier.
 
Recently, F.~Husze\'{a}r~\cite{Ferenc} offers an information-theoretic interpretation to IRM and suggests that the regularization term $||\nabla_{g|g=1} R^{e}(g\circ \Phi)||_{2}^{2}$ in Eq.~(\ref{eqIRM}) is closely related to a conditional mutual information constraint $I(y;e|\Phi(\mathbf{x}))$, where $e$ is the environment index. In this case, $y$ does not bring any new information to infer the environment index given $\Phi(\mathbf{x})$. Following this interpretation, Li \emph{et al}.~\cite{li2022invariant} develop the Invariant Information Bottleneck (IIB) by optimizing the following objective:
\begin{equation}\label{eq222}\small
\text{max}_{\Phi}\{I[y; \Phi(\mathbf{x})]-\lambda_{1} I(\mathbf{x};\Phi(\mathbf{x}))-\lambda_{2}I(y;e|\Phi(\mathbf{x}))\}.
\end{equation}

However, IRM and its variants typically assume explicit environment indices are given~\cite{creager2021environment,liao2022decorr}, whereas in practice and also our FGVC data, we only have a single training set in which no environmental partition is available. 
On the other hand, most of the studies on IRM just evaluate on MNIST-like datasets. In this sense, it is uncertain about the practical performance of IRM in challenging computer vision tasks like FGVC on large-scale image data. 

\section{The Proposed Method}
\begin{figure*}[hbpt]
  \begin{center}
  \includegraphics[width=0.99\textwidth]{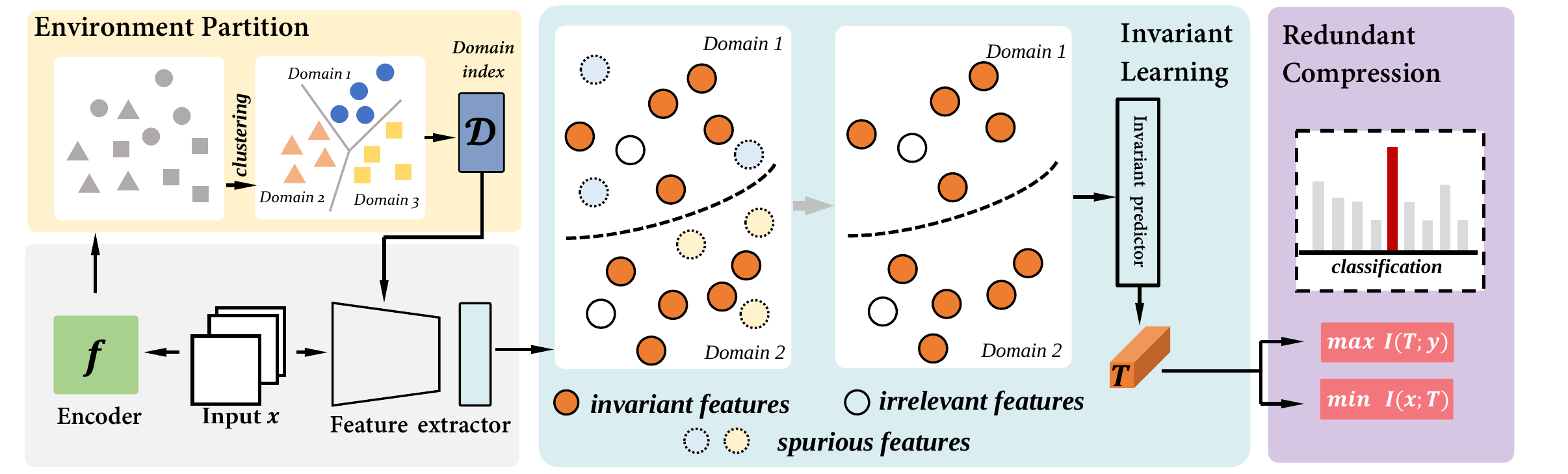}
  \end{center}
	\setlength{\abovecaptionskip}{-0.2cm} 
\vspace{-0.3cm} 
  \caption{Overview of the proposed method. Our framework first divides the training data into different environments by performing soft $k$-means clustering in the latent space. Then, we apply IRM and IB in a joint objective to ensure the invariance and minimum sufficiency of the learned representation.}
  \label{Overview}
\vspace{-0.5cm} 
\end{figure*}

In this section, we describe our IMS. As shown in Fig.~\ref{Overview}, it consists of three core components, including environment partition, invariant representation learning module, and redundancy reduction module. 

\subsection{Environment Partition of IRM For Invariant Representation Learning}
As mentioned above, IRM and its variants assume the partition of environments is predefined. For example, in the problem of objective recognition on PACS datasets~\cite{li2017deeper}, people usually train on samples from the ``photo", ``art", and ``cartoon" domains, and test on ``sketch" domain.
Unfortunately, for FGVC data, we only have a single environment, which means that samples collected from multiple locations or over different periods of time are merged together. 
Therefore, a direct application of IRM is infeasible. 
To implement our IMS, we first need to determine a reliable environment partition of the training set.
In this work, we explore the soft $k$-means clustering approach on the features of the last layer of ResNet-50 to achieve this goal.
The optimization objective is defined as follows:
\begin{equation}\small
E = \sum_{l=1}^{k} \sum_{i=1} ^{n} \gamma_{il} ||\Phi(\mathbf{x})-\mu_{l}||^{2} ,
\label{soft}
\end{equation}
where $k$ is the number of environments, $n$ is the number of instances.
$\mu_{l}$ represents the corresponding center point of the cluster.
$\gamma_{il}$ can be interpreted as the probability of $\mathbf{x}$ belonging to domian $l$, and for each $\mathbf{x}$ satisfies $\sum_{l=1}^{k} \gamma_{il} =1$.
We then take the instances of the acquired environment label into the loss function design.
%and formulate our objective as:
%\begin{equation}\small
% \text{max}_{\Phi}\{I[y; \Phi(\mathbf{x})]-\beta I(y;e|\Phi(\mathbf{x})\},
%\end{equation}
%where $e$ is the environment index.

Apart from the soft k-means, we also tested the Decorr~\cite{liao2022decorr} method, which is the SOTA environment partitioning approach. However, when the number of environments is small, our results are very close. When the number of environments is large, Doccer does not show any performance improvement or even performs worse. One possible reason for this is that this division method leads to an insufficient number of images in each environment. (Specific results can be found in Appendix B.)

\subsection{IB for Redundant Feature Compression}

The IB principle prescribes to learn a compressed representation~\cite{tishby2000information, tishby2015deep}, which can also be interpreted as an approximation to the minimum sufficient statistics~\cite{shamir2010learning,achille2018emergence}. Given input variable $\mathbf{x}$ and desired response $y$ (e.g., class labels), the IB approach aims to extract from $\mathbf{x}$ a compressed representation $\Phi(\mathbf{x})$ that is most informative to predict $y$. 
In this process, mutual information $I(\mathbf{x};\Phi(\mathbf{x}))$ is used to measure information compression, whereas the mutual information $I(y;\Phi(\mathbf{x}))$ is used to quantify the predictive ability of $\Phi(\mathbf{x})$ to $y$.
Formally, this objective is formulated as finding $\Phi(\mathbf{x})$ such that the $I(y;\Phi(\mathbf{x}))$ is maximized, while keeping $I(\mathbf{x};\Phi(\mathbf{x}))$ below a threshold $\alpha$: 
\begin{equation}\label{eq:IB_original}\small 
    \argmax_{\Phi(\mathbf{x})\in\Delta} I(y;\Phi(\mathbf{x}))\quad \text{s.t. } \quad I(\mathbf{x};\Phi(\mathbf{x}))\leq \alpha, 
\end{equation} 
where $\Delta$ is the set of random variables $\Phi(\mathbf{x})$ that obey the Markov chain $y - x - \Phi(\mathbf{x})$. In practice, it is hard to solve the above constrained optimization problem of Eq.~(\ref{eq:IB_original}), and $\Phi(x_{i})$ is found by maximizing the following IB Lagrangian: 
\begin{equation}\label{eq:IB_Lagrangian} 
  \mathcal{L}_{\text{IB}} = I(y;\Phi(\mathbf{x}))-\beta I(\mathbf{x};\Phi(\mathbf{x})), 
\end{equation} 
where $\beta$ is a Lagrange multiplier that controls the trade-off between the performance of $\Phi(\mathbf{x})$ on task $y$ (as quantified by $I(y;\Phi(\mathbf{x}))$) and the complexity of $\Phi(\mathbf{x})$ (as measured by $I(\mathbf{x};\Phi(\mathbf{x}))$). 
We use IB approach to remove these redundancy features by the objective of $\max I(y;\Phi(\mathbf{x}))-\beta I(\mathbf{x};\Phi(\mathbf{x}))$. 
There are different ways to parameterize IB by neural networks. In general, the maximization of $I(y;\Phi(\mathbf{x}))$ is equivalent to the minimization of CE loss, which turns the objective of deep IB into a standard CE loss regularized by a differentiable mutual information term $I(\mathbf{x};\Phi(\mathbf{x}))$. 
If representation $\Phi(\mathbf{x})$ is a deterministic transformation of $\mathbf{x}$ such that the conditional entropy $H(\Phi(\mathbf{x})|\mathbf{x})$ reduces to zero (because there is no uncertainty on the mapping), then we have $I(\mathbf{x};\Phi(\mathbf{x})) = H(\Phi(\mathbf{x})) - H(\Phi(\mathbf{x})|\mathbf{x}) = H(\Phi(\mathbf{x}))$~\cite{strouse2017deterministic,ahuja2021invariance}. Thus, our final objective reduces to Eq.~(\ref{eqIB}), in which $H$ denotes entropy:
\begin{equation}\label{eqIB}
  \mathcal{L}_{\text{IB}}= \mathcal{L}_{\text{CE}}+\beta H(\Phi(\mathbf{x})).
\end{equation}

\subsection{Joint Training of IRM and IB}
To guarantee the invariance and the minimum sufficiency of learned representations, we combine the merits of both IRM (i.e., Eq.~(\ref{eqIRM})) and IB (i.e., Eq~(\ref{eqIB})) and propose the following joint learning objective:
\begin{equation}\label{eqtotalIMS}\small
  \mathcal{L}_{\text{IMS}}= \sum _{i=1}^{K} \mathcal{L}_{\text{CE}}+\eta ||\nabla_{g|g=1} R^{e}(g\circ \Phi)||_{2}^{2} + \beta H(\Phi(\mathbf{x})). 
\end{equation}
Eq.~(\ref{eqtotalIMS}) involves an entropy minimization regularization term, which is hard to control in practice due to the difficulty of entropy estimation in high-dimensional space. For example, the class token of ViT is usually $768$-dimensional, the last hidden layer of ResNet is $2,048$-dimensional. To this end, we introduce the recently proposed matrix-based R{\'e}nyi' $\alpha$-order entropy functional~\cite{giraldo2014measures} to evaluate $H(\Phi(\mathbf{x}))$ and set $\alpha=1.01$ to approximate Shannon entropy. We directly give the definition as below.

{\bf Definition 1.} Let $\kappa : \mathcal{X} \times \mathcal{X} \mapsto \mathbb{R}$ be a real valued positive definite kernel that is also infinitely divisible~\cite{bhatia2006infinitely}. 
Given $X = \{x_{1}, x_{2}, ..., x_{n}\}$ and the Gram matrix $K$ obtained from evaluating a positive definite kernel $\kappa$ on all pairs of exemplars, that is $(K)_{ij}=\kappa(x_{i},x_{j})$, a matrix-based analogue to R{\'e}nyi's $\alpha$-order entropy for a normalized positive definite~(NPD) matrix $A$ of size $n\times n$, such that tr($A$) = 1, can be given by the following functional:
\begin{equation}\small
{\bf S}_{\alpha}(A)=\frac{1}{1-\alpha}\log_2 \left(\tr (A^{\alpha})\right)=\frac{1}{1-\alpha}\log_{2}\left(\sum_{i=1}^{n}\lambda _{i}(A)^{\alpha}\right),
\end{equation}
where $A_{ij} = \frac{1}{n} \frac{K_{ij}}{\sqrt {K_{ii}K_{jj}}}$ and $\lambda_{i}(A)$ denotes the $i$-th eigenvalue of $A$.

This new way of estimation avoids density estimation and the variational lower bound approximation (by an extra neural network), which makes implementation extremely simple.
Our generalization scheme in Eq.~(\ref{eqtotal}) is most similarly to~\cite{ahuja2021invariance, li2022invariant}. Different from ours, \cite{ahuja2021invariance} assumes that environmental partition is available and does not consider FGVC task. Moreover, \cite{ahuja2021invariance} makes a Gaussian assumption on $\Phi(\mathbf{x})$ and uses the variance instead of the differential entropy for simplicity. The final objective in \cite{ahuja2021invariance} is given by:
\begin{equation}\label{eqAll}\small
\vspace{-0.1cm} 
  \mathcal{L}= \sum _{i=1}^{K} \mathcal{L}_{\text{CE}}+\eta ||\nabla_{g|g=1} R^{e}(g\circ \Phi)||_{2}^{2} + \beta \text Var\Phi(\mathbf{x}). 
%\vspace{-0.2cm} 
\end{equation}

\begin{figure}[htbp]
  \begin{center}
  \includegraphics[width=0.87\textwidth]{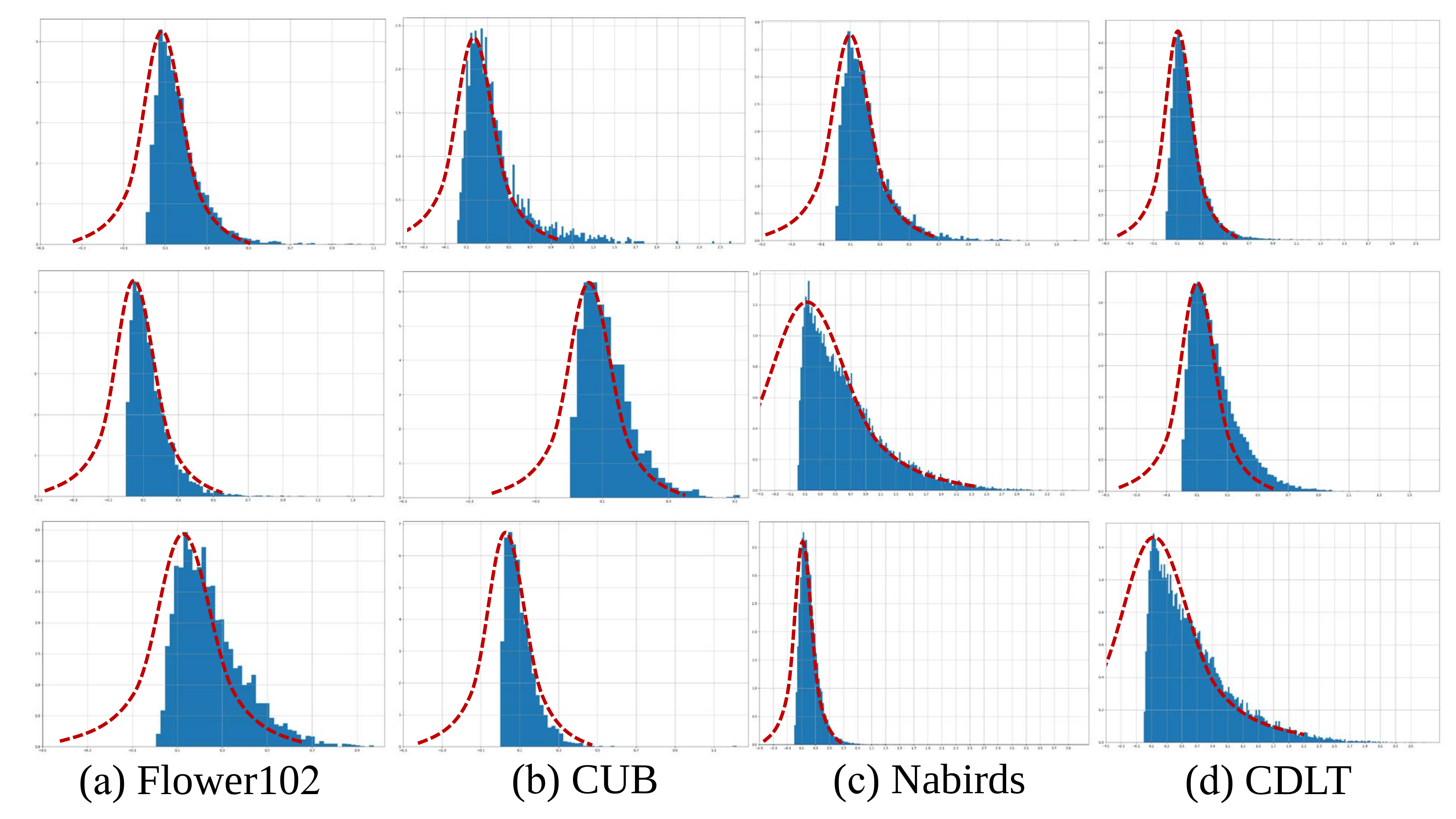}
  \end{center}
	\setlength{\abovecaptionskip}{-0.1cm} 
\vspace{-0.2cm} 
\caption{Feature distribution visualization. The pre-trained ResNet-50 is used to extract the features from the training set. The red dotted line shows the Gaussian distribution for the current spindle.}
\label{FRA}
\vspace{-0.2cm} 
\end{figure}

However, Gaussian assumption does not hold for FGVC data. To illustrate this problem, we visualized the feature distribution at the last layer in four datasets, and the results are shown in Fig.~\ref{FRA}. We randomly selected a position from the $2,048$ latitude feature and extracted the values of the following five consecutive latitudes for statistics. Obviously, it was a right-skewed distribution instead of Gaussian distribution.
Our IMS addresses both limitations by soft $K$-means clustering on latent representations to partition environments, and by the recently proposed matrix-based R{\'e}nyi's $\alpha$-order entropy~\cite{giraldo2014measures} to estimate the entropy term without any distributional assumption.

In~\cite{li2022invariant}, the authors replace the third term in Eq.~(\ref{eqAll}) with a conditional mutual information regularization $I(y;e|\Phi(x))$ and approximate this term as the difference of two CE losses as shown in Eq.~(\ref{eqtotal}).
%In~\cite{li2022invariant}, the authors replace the second term in Eq.~(\ref{eqtotal}) into the two different CE losses by variational approximation~\cite{farnia2016minimax}. 
However, it does not make explicit estimation but uses the approximation ability of neural networks and simply subtracts these two losses, which makes the approximate error cannot be estimated. 
Instead, our method is a much more accurate estimation to both mutual information and conditional mutual information terms.
%\begin{equation}\label{AAAI}
%\begin{aligned}
%    I[y;e|\Phi(\mathbf{x})] & = H(y|\Phi(\mathbf{x}))- H(y|e,\Phi(\mathbf{x})). \\
%\end{aligned}
%\end{equation}

\begin{equation}\label{eqtotal}
\begin{aligned}
I(y; e|\Phi(\mathbf{x})) &= H(y|\Phi(\mathbf{x})) - H(y|e,\Phi(\mathbf{x})) \\
          &= \text {min}_{f_{i},\Phi}\mathbb{E}_{x,y}[L(y,f_{i}(\mathbf{x}))] \\
        &  - \text {min}_{f_{i},\Phi}\mathbb{E}_{x,y,e}[L(y,f_{e}(\mathbf{x}),e))],
\end{aligned}
\end{equation}
where $f_{i}$ and $f_{e}$ are classifiers, $f_{i}$ take the feature $\Phi(\mathbf{x})$ as the input, $f_{e}$ take the feature $\Phi(\mathbf{x})$ and the index of the environment $e$ as the input. 

\section{Experiments}
\subsection{Datasets and Implementation Details}
We conducted experiments on seven benchmark FGVC datasets with fixed train and test splits, as well as two additional datasets with artificially introduced distributional shifts. 
The detailed statistics can be found in Table~\ref{Summary}.

\begin{table}[hbpt]\small
\setlength{\abovecaptionskip}{-0.15cm}
\caption{Summary statistics of fine-grained datasets.}
 \label{Summary}
  \begin{center}
  \begin{tabular}{c|c|c|c}
  \hline
  Dataset & Category & Training & Testing \\
  \hline
Flower-102~\cite{nilsback2008automated}      & 102   & 2066   & 6123  \\ 
CUB-200-2011~\cite{wah2011caltech}           & 200   & 5994   & 5794  \\ %&$\times$\\
Nabirds~\cite{van2015building}               & 550   & 23929  & 24633 \\ %&$\times$ \\
CDLT~\cite{ye2023cdlt}                       & 258   & 5091   & 4458  \\ %&$\times$\\
Stanford Cars~\cite{krause20133d}            & 196   & 8144   & 8041  \\
Stanford Dog~\cite{khosla2011novel}          & 120   & 12000  & 8580  \\ %&$\times$\\
Aircraft~\cite{maji2013fine}                 & 100   & 6667   & 3333  \\
\hline
Flower-shift                                 & 102   & 2066   & 6123  \\ 
CDLT-cd                                    & 258   & 5091   & 4458  \\ %&\checkmark \\
  %&\checkmark \\ 
  \hline
  \end{tabular} 
 \end{center}  
  \vspace{-0.6cm} 
\end{table} 

CDLT is our self-constructed dataset, which includes images of natural contexts recorded over 47 consecutive months and covers more than 250 subclasses. Each image has the corresponding season information (Examples from CDLT can be found in Appendix C) 
\footnote{We will make this dataset publicly available. Currently, a portion of CDLT can be downloaded from 
\url{https://drive.google.com/drive/folders/1LuWyQq74ZRe1Zo-Nvg3U2gUB5NRqLbX3.}}.
CDLT-cd is a reconstructed dataset derived from CDLT, where we grouped winter and spring as S1, and summer and autumn as S2 based on the growth of instances.      
For each subclass in the training set, 75\% of the data are randomly sampled from S1 and the remaining 25\% from S2.      
Flower-shift is reconstructed from Flower102.  The dataset is divided into two domains based on flower color or growth cycle, with 75\% of the training data sampled from one domain and the remaining 25\% used for testing.  Following the partition ratio of the Flower102 dataset, the number of training and test images is 2,108 and 6,081, respectively.   

In our experiments, pre-trained ViT-B-16~\cite{dosovitskiy2020image}, ResNet-50~\cite{he2016deep}, and VGG-16~\cite{simonyan2014very} are used as backbones, and the input image size is $448\times448$. 
Our pre-processing follows the popular configurations~\cite{he2022transfg,wang2021feature}. 
Specifically, we use data augmentations, including random cropping and horizontal flipping in the training procedure. 
During inference, only center cropping is used.
All datasets are trained using stochastic gradient descent (SGD) with a momentum of 0.9. The learning rate is set to 3e-2, and a cosine decay function is applied to an optimizer step.
Our implementation is based on PyTorch and utilizes four NVIDIA A6000 GPUs. The evaluation metric for all experiments is top-1 accuracy.

\subsection{Spurious Correlation Leads to Misleading Classification}
\begin{figure}[htbp]
  \begin{center}
  \includegraphics[width=0.87\textwidth]{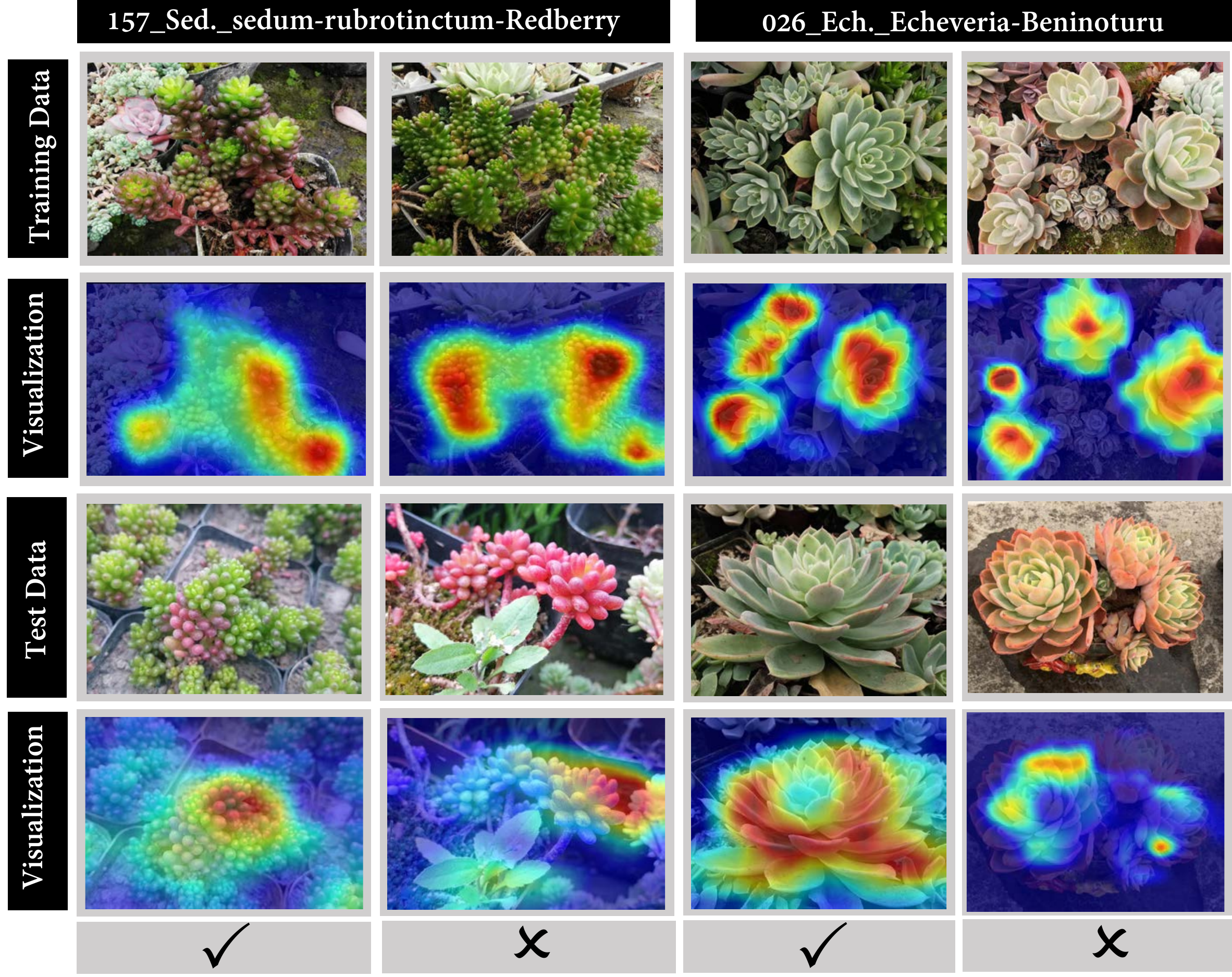}
  \end{center}
	\setlength{\abovecaptionskip}{-0.1cm} 
\vspace{-0.4cm}
  \caption{Grad-CAM~\cite{selvaraju2017grad} visualization for the last convolutional layer of the ResNet-50 on the CDLT dataset. \checkmark~indicates correct classification, whereas~\XSolidBrush~indicates wrong decision.} 
  \label{Spurious_Correlation}
\vspace{-0.2cm}
\end{figure}

We demonstrate the training and test set in Fig.~\ref{Spurious_Correlation} and visualize the attention regions during model classification.
In the first subclass~(e.g., \textit{Sed.rub.Redberry}), the majority of training data has green tops. 
As for the test set, it includes instances similar to the training set, but some have significant differences~(the instance color changed from green to red).
The model accurately identifies the crucial region and makes the correct decision for similar instances. However, for dissimilar instances within the same subclass, the model focuses mainly on the background and irrelevant green leaves, resulting in an obvious mistake.
The second subclass~(e.g., \textit{Ech.Beninoturu}) consists of green or light yellow training instances. For similar instances in the test set, the model focuses on the whole instance.
However, when there is a variation in color, the model only focuses on regions with analogous colors present in the training set.
These examples illustrate that the model often takes shortcuts when making decisions, such as relying heavily on color.
In contrast, experienced experts generally avoid interference from the spurious correlations between colors and labels because they are aware that instance colors typically change over time and are therefore not always reliable.

\subsection{Model Configuration}
Model configuration experiments are conducted on the CUB and CDLT-cd datasets to verify the validity of the individual components and to determine the hyperparameters. 

\textbf{Invariant Representation Learning ($\boldsymbol{\eta}$):}
To verify the effectiveness of IRM and investigate the influence of the parameter $\eta$, extensive experiments are carried out, and the results are presented in Table~\ref{IRM}. 
To maintain consistency in the experimental environment, we divide the training set into five environments, i.e., $K=5$. $Batch size$ is abbreviated as `Bs' and set to be 24.
As expected, when $\eta$ is set properly (e.g., $\eta \in [0.001~0.005]$), IRM can alleviate the influence of distributional shift to a certain extent and prompt the invariant representation learning capability.
However, if we further increase $\eta$, our performance suffers from a slight drop (but is still better than ERM, i.e., $\eta=0$). This result suggests that $\eta=0.005$ could be a reliable choice for our IMS.

\begin{table}[h]\footnotesize
\setlength{\abovecaptionskip}{-0.15cm}
 \begin{center}
\caption{Experimental results using varied $\eta$.}
\label{IRM}
 \begin{tabular}{c|ccccc|c}  
 \bottomrule 
  $\eta$  & 0.001 & 0.005 & 0.01 & 0.05 & 0.1 & ERM   \\
 \hline
  CUB     & 90.79  & \bf 91.13  & 90.96 & 90.74 & 90.72  & 90.41 \\
  CDLT-cd & 69.07  & \bf 69.33  & 68.85 & 69.05 & 68.49  & 68.47 \\
  \hline  
\end{tabular} \end{center}
\vspace{-0.6cm}  
\end{table}

\textbf{Redundancy Compression ($\boldsymbol{\beta}$):}
To investigate the effect of the redundancy reduction term $H(\Phi(\mathbf{x}))$, we conduct experiments with different values of $\beta$, and the results are given in Table~\ref{IB}. 
As expected, information compression enhances the network representation capability, which also suggests that a large amount of redundant features are learned by the model during the training process. 

\begin{table}[hbpt]\footnotesize
\setlength{\abovecaptionskip}{-0.04cm}
\caption{Experimental results using varied $\beta$.}
\vspace{-0.3cm}  
\label{IB}
 \begin{center}
 \begin{tabular}{c|cc cccc|c}  
 \bottomrule 
  $\beta$   & 0.0005 & 0.001 & 0.005 & 0.01 & 0.05 &0.1 & ERM\\
 \hline
  CUB      & 90.59 &  90.66 & 90.72 & 90.78 & \bf 90.84 & 90.67 & 90.41\\
  CDLT-cd  & 68.73 &  68.84 & 68.88 & 69.27 & \bf 70.01 & 68.93 & 68.43 \\
  \hline  
\end{tabular} \end{center}
\vspace{-0.4cm}    
\end{table}

Besides, we observe that increasing $\beta$ leads to higher accuracy.
However, the performance slightly drops when the balance parameter $\beta$ increases from 0.05 to 0.1, which infers that when $\beta$ is equal to 0.05, the redundant information contained in the target has been squeezed out.
Therefore, we chose $\eta = 0.005$ and $\beta = 0.05$ for all subsequent experiments as it ideally balances the computational complexity and accuracy.

\textbf{Number of Environments $K$:}
To explore the impact of the number of environments on IRM learning, we cluster the training data into several environments and report the results in Table~\ref{Num_Domains}. 
\begin{table}[h]\footnotesize
\setlength{\abovecaptionskip}{-0.04cm}
 \setlength{\tabcolsep}{7pt}
 \begin{center}
\caption{Different values of $K$ and batchsize~(Bs).}
\label{Num_Domains}
 \begin{tabular}{c|c| ccccc}  
\hline
  $ $   & ERM & $K$=2 & $K$=3 & $K$=4 & $K$=5 & $K$=6    \\
   \hline
  Bs=8  & 90.33  &\bf90.75 & 90.51 & 90.54  & 90.59     & 90.66  \\
  Bs=16 & 90.38  &  90.66  & 90.62 & 90.81  & \bf 90.96 & 90.71  \\
  Bs=24 & 90.41  &  90.89  & 90.73 & 90.84  & \bf 91.13 & 90.89  \\
  Bs=32 & 90.40  &  91.00  & 90.76 & 90.94  & \bf 91.27 & 90.90  \\
  Bs=40 & 90.40  &  90.92  & 90.84 & 91.11  & \bf 91.34 & 91.09  \\
 \hline 
  Random&   -    &  90.57  & 90.50 & 90.73     & 90.88     & 90.80 \\
  \hline  
\end{tabular} \end{center}
\vspace{-0.4cm}  
\end{table}
It is evident from the results that the batch size has a negligible effect on the learning performance of ERM within the current range, whereas it does influence the performances of IRM.
As the batch size increases, the accuracy of IRM always improves accordingly. 
This make sense, since an increased number of samples in the mini-batch enhances the reliability~(and precision) of our entropy estimator.
When the number of environments increases, the accuracy of the model begins to increase because more intra-class differences can be readily expressed, which reduces the influence of spurious features on discrimination.
Increasing the number of domains will result in fewer samples per environment, which may lead to overfitting or poor generalization performance.
For the following experiments, we set $K$=5.

\subsection{Performance Evaluation}
The experimental results and analysis of IMS compared with recent SOTA methods on seven datasets are presented in Table~\ref{CDLT_result} and Table~\ref{resul2}.
In particular, the pre-trained model used in Sim-trans is based on ImageNet 21k, while our method and most of the compared models utilize ImageNet 1k. To ensure a fair comparison, we re-ran their experiments using ImageNet 1k as the backbone model.

\begin{table}[hbpt]\small
  \centering
 \caption{Comparisons with the SOTA FGVC methods. ``Ori." refers to CDLT and Flower-102. ``Shift" refers to CDLT-cd and Flower-shift, respectively.}
%\vspace{-0.2cm} 
  \label{CDLT_result}
  \begin{tabular}{l |c | cc|cc}
    \toprule
\multirow{2}{1.3cm}{\textbf{Method}}    &\multirow{2}{1.5cm}{\textbf{Backbone}} %&\multirow{2}{2.3cm}{\textbf{CUB-200-2011}} 
    &\multicolumn{2}{c|}{\textbf{CDLT}} & \multicolumn{2}{c}{\textbf{Flower}} \\ %& \multirow{2}{1cm}{\textbf{Year}} \\
    \cline{3-6}
   & & \textbf{Ori.} & \textbf{Shift}  & \textbf{Ori.} & \textbf{Shift} \\
\hline
%NTS-Net\cite{Yang2018Learning}  & VGG-16        & 59.1 & 54.9        &  -    &  -        \\  
%MCL\cite{chang2020devil}        & VGG-16        & 62.9 & 59.2        &  -    &  -        \\
%MOMN\cite{min2020multi}         & VGG-16        & 55.6 & 48.9        &  -    &  -        \\
NTS-Net\cite{Yang2018Learning}~(ECCV,2018)    & ResNet-50   & 74.3  & 69.0  & 96.8  & 82.9 \\  
MCL\cite{chang2020devil}~(TIP,2020)           & ResNet-50   & 71.8  & 67.1  & 94.1  & 84.3 \\
MOMN\cite{min2020multi}~(TIP,2020)            & ResNet-50   & 60.7  & 56.6  & 96.2  & 82.6 \\
PMG~\cite{du2021progressive}~(TPAMI,2021)     & ResNet-50   & 70.2  & 68.4  & 95.1  & 83.8 \\

ViT~\cite{dosovitskiy2020image}~(2020)        & ViT-B-16    & 76.0  & 68.4  & 99.3  & 98.6 \\
TransFG\cite{he2022transfg}~(AAAI,2022)       & ViT-B-16    & 76.6  & 69.5  & 99.4  & 98.8 \\	
FFVT\cite{wang2021feature}~(2021)             & ViT-B-16    & 75.6  & 69.6  & 99.4  & 98.3 \\
IRM~\cite{arjovsky2019invariant}~(2019)       & ViT-B-16    & 77.0  & 69.3  & 99.3  & 98.8 \\
IB-IRM~\cite{ahuja2021invariance}~(NIPS,2021) & ViT-B-16    & 77.2  & 69.4  & 99.4  & 98.8 \\
\hline
IMS	  & ResNet-50    & 72.7    &\bf 69.6    &  96.2    & \bf 86.1     \\
%	IMS(k-means)                & ViT-B-16    & \bf77.9    &\bf 70.4    & \bf 99.4     & \bf 98.9 \\
IMS   & ViT-B-16    & \bf 77.9      &\bf 70.8    & \bf99.4            & \bf 99.1 \\
\bottomrule
  \end{tabular}
\vspace{-0.2cm} 
\end{table}

ResNet-50 is found to generally outperform VGG-16 among convolutional neural network~(CNN)-based models.
The best performance among CNN-based models is achieved by NTS-Net, which uses multiple sub-networks for feature extraction at different granularities, resulting in rich feature information.
However, NTS-Net requires a complex training process with several loss constraints at different stages.
Another noteworthy method is PMG~\cite{du2021progressive}, which has achieved competitive results on CUB. However, it requires that the distinctive features of each instance should be visible, but this is difficult to achieve in a complicated environment.
The performance of ViT-based models is superior to that of CNN-based models.
This may owe to the inherent self-attention mechanism of ViT, which enables the model better focus on the discriminative regions of the instance.
However, this advantage is based on the assumption that the features learned in the training set remain stable discriminative in the test set. 
When some features change over time and result in distributional shifts, the interaction mechanism may aggravate the redundancy of information. 
Comparing the results of ResNet50-based models and ViT-based models, it can be observed that the impact of the distributional shift on FGVC is universal, especially for ViT-based models, where the decline is 0.2\% higher than that of ResNet50-based models.

\begin{table}[hbpt]\footnotesize
  \centering
 \caption{Comparisons with the SOTA FGVC methods on CUB, Nabirds, Stanford Cars, Stanford Dogs, and Aircraft datasets.}
%\vspace{-0.4cm} 
  \label{resul2}
 \setlength{\tabcolsep}{4pt}
 \begin{tabular}{l|c|c|c|c|c|c}  
\toprule   
  \textbf{Method}  &\textbf{Backbone}   &\textbf{CUB}  & \textbf{Nabirds}  & \textbf{Car} & \textbf{Dog} & \textbf{Air}\\  
\hline
%B-CNN~\cite{zhu2017b}                        & VGG-16     &84.1   & 79.4     \\
%PMG~\cite{du2021progressive}~(TPAMI,2021)    & VGG-16     &89.0   & 87.0     \\
M2DRL~\cite{he2019and}~(IJCV,2019)           & VGG-16     & 87.2  & -     & 93.2    &-   &- \\
MGE-CNN~\cite{zhang2019learning}~(ICCV,2019)  & ResNet-50  & 88.5  & 88.6  &93.4  &-      &-\\
%CS-Parts~\cite{korsch2019classification}     & ResNet-50  & 89.5  & 88.5   \\	
API-Net~\cite{zhuang2020learning}~(AAAI,2020) & ResNet-101 & 87.7  & 86.2  &94.9  &-      & 93.4 \\
MCL~\cite{chang2020devil}~(TIP,2020)           & ResNet-50 & 87.3  & -     &93.7  & -  & 92.5 \\
PMG~\cite{du2021progressive}~(TPAMI,2021)     & ResNet-101 & 90.0  & 88.4  &95.1  &-      & 93.4 \\
ViT~\cite{dosovitskiy2020image}~(2020)        & ViT-B-16   & 90.4  & 89.6  &93.7  &92.0   & 92.2 \\
IRM~\cite{arjovsky2019invariant}~(2019)       & ViT-B-16   & 91.1  & 89.7  &93.4  & 92.1  & 91.8 \\
IB-IRM~\cite{ahuja2021invariance}~(NIPS,2021) & ViT-B-16   & 91.1  & 89.7  &94.0  & 92.4  & 92.8\\
Rams-Trans~\cite{hu2021rams}~(ACM MM,2021)    & ViT-B-16   & 91.3  & -     &-     & 92.4  & - \\
TPSKG~\cite{liu2022transformer}~(Neurocomputing,2022) & ViT-B-16   & 91.3  & 90.1 &-  &92.5 &-           \\
P2P-Net~\cite{yang2022fine}~(CVPR,2022)       & ResNet-50  & 90.2  & -     &\bf95.4  &-  &94.2 \\
CHRF~\cite{liu2022focus}~(ECCV,2022)          & ResNet-101 & 90.8  & -     &\bf95.4  &-  &\bf94.7 \\
SIM-Trans~\cite{sun2022sim}~(ACM MM,2022)     & ViT-B-16   & 91.5  & -     &-     & -  & - \\
%RCE~\cite{tang2023weakly}~(CVPR,2023)         & SwinTrans  & 91.2  & - \\
\hline
%IMS(k-means)                                 & ViT-B-16   &\bf91.4&\bf 90.0   \\ 
    IMS                                       & ViT-B-16   &\bf91.6&\bf 90.2  & 94.6  & \bf92.8 & 93.1 \\ 
\bottomrule
  \end{tabular}
\vspace{-0.4cm} 
\end{table}

IRM~\cite{arjovsky2019invariant} is one of the most classical approaches to addressing distributional shift problems.
It can be observed that it achieves a performance improvement of 0.9\% (CDLT-cd) when dealing with data with distributional shift. Furthermore, it also demonstrates a maximum performance improvement of 0.7\% (CUB) on regular FGVC datasets. One significant reason for this improvement is that the model can comprehend instances in diverse environments, thereby learning more discriminative features.
On the other hand, IB-IRM~\cite{ahuja2021invariance} initiated the idea of combining IB and IRM. IB-IRM has shown improved recognition accuracy compared to ViT. 
However, we observed that this improvement is rather similar or almost the same to the result which uses IRM alone.
Thus, the entropy regularization term $\text{Var}~\Phi(\mathbf{x})$ proposed by IB-IRM does not really contribute to the performance gain, which further suggests that the Gaussian assumption made by IB-IRM does not hold for FGVC data.
In comparison, on the CDLT-cd, IMS achieves a performance gain of 1.5\% based on IRM. It also achieves a general performance improvement ranging from 0.5\% to 1.3\% on typical FGVC datasets. This suggests that removing redundant information can better assist model learning. 
Our IB can learn the minimal sufficient representation of features from diverse environments without any additional assumptions or approximations, which explains why it achieves the SOTA performance.
It is important to emphasize that our method was specifically developed to tackle the challenges of OOD scenarios in FGVC. However, it is apparent that the OOD problem in the Stanford Cars and Aircraft datasets is not pronounced~\cite{ye2023image}. Nonetheless, our method still achieved competitive results on these datasets, which validates the effectiveness and generalizability of the proposed approach.

Further, we conducted an experiment to evaluate the portability of our approach. We combined our method with the current state-of-the-art methods, and the results were reported in Table~\ref{combined}.

\begin{table}[hbpt]\small
\centering
\caption{Performance combined with the current state-of-the-art methods. 
The values in parentheses represent the gains our algorithm achieved over the original model.}
%\vspace{-0.4cm} 
  \label{combined}
 \begin{tabular}{c| c|c|c}  
 \bottomrule 
  	 Method   & Backbone  & CUB & CDLT-cd \\  
\hline  
SPS~\cite{huang2021stochastic}  &ResNet-50 & 87.3 (+0.4)  & 68.5(+1.1)  \\
FFVT~\cite{wang2021feature}     &ViT-B-16  & 91.6 (+0.1)  & 69.6(+0.6)\\
  \hline  
\end{tabular}
\vspace{-0.2cm} 
\end{table}

It can be observed that our method contributes to further enhancing the performance of some state-of-the-art approaches. For instance, SPS achieved a performance improvement of 0.4\% on the CUB dataset, and one possible reason could be the effective reduction of feature redundancy achieved by IB.
The improvements we obtained were even more significant on datasets with distributional shift, where SPS achieved a performance gain of 1.1\%. This highlights the impact of distribution drift on FGVC tasks and suggests a direction for future algorithmic enhancements.
Additionally, we also evaluated the model's performance on different forms of data. More analysis can be found in Appendix D.

\subsection{Compare with Data Augmentation Methods}
Data augmentation typically enhances the generalization capability of models~\cite{guo2023zero,wang2023learning}, but it is not always the case in FGVC tasks. 
An important reason is that when data augmentation methods increase the number of images, they also introduce numerous non-discriminative regions.
This is not particularly helpful for FGVC tasks that rely on subtle local differences for classification.
To illustrate this issue, we conducted experiments using three representative data augmentation methods: Cutout~\cite{devries2017improved}, Mixup~\cite{zhang2017mixup}, and CutMix~\cite{yun2019cutmix}. 
We followed the settings outlined in their respective papers to ensure consistency and accuracy in our results. 
We trained the backbone network on the CUB dataset as the ``Baseline''. The results are reported in Table~\ref{comparison_AU}.

\begin{table}[hbpt]\small
\centering
\caption{Performance comparison of data augmentation methods.}
%\vspace{-0.4cm} 
  \label{comparison_AU}
 \begin{tabular}{c| c|c|c}  
 \bottomrule 
  	 Method    & ViT-B-16 & ResNet-50 & VGG-16 \\  
    \hline
     Baseline      & 90.4  & 85.5 &  78.0  \\
    \hline
Cutout         &90.1(-0.3)  & 83.5(-2.0)   &78.2(+0.2)\\
Mixup         &90.6(+0.2)  &85.9(+0.4)   &78.0(-0.0)\\
CutMix        &91.0(+0.6)  &86.2(+0.7)   &77.3(-0.7)\\
  \hline  
\end{tabular}
\vspace{-0.2cm} 
\end{table}

It is evident that although CutOut increases the number of images in the CUB dataset, it does not provide the expected gains to the model. On the contrary, it results in performance decay in ViT and ResNet-50, with ResNet-50 experiencing a 2\% loss in performance.  
Mixup prevents performance decay, but the gains achieved are not ideal.  When VGG-16 is used as the backbone network, the model does not receive any benefit at all. While CutMix improves the performance of ViT and ResNet-50, it significantly deteriorates VGG-16's performance~(by 0.7\%). The experimental results confirm the point made at the beginning of this section that using data augmentation methods directly does not improve the model's generalization. In fact, several recent studies have implicitly suggested this viewpoint~\cite{huang2021snapmix,li2020attribute}.

\subsection{Ablation Study}
To evaluate the effectiveness and generalization ability of the proposed IMS, we conducted ablation experiments on CDLT-cd datasets. 
First, we evaluated the performance of the basic model in FGVC tasks by implementing a ``Baseline" model that learned fine-grained objectives without taking any action. 
Second, we evaluated the effectiveness of the IRM approach by implementing a ``w/o IB" baseline that learned fine-grained objectives without using the redundant information compression action. 
Then, we evaluated the effectiveness of our proposed IB loss by implementing a ``w/o IRM" baseline to compare the results with and without adding IRM to the loss function. 
%The training process for these methods is illustrated in Fig.~\ref{CDLT_Acc}.
%\begin{figure}[htbp]
%  \begin{center}
%  \includegraphics[width=0.42\textwidth]{fig/CUB_Training_process.pdf}
%  \end{center}
%	\setlength{\abovecaptionskip}{-0.5cm} 
%\vspace{-0.4cm}
%  \caption{Visualization of the training process. Horizontal and vertical axes indicate the training epoch and Acc. respectively.}
%  \label{CDLT_Acc}
%\vspace{-0.2cm} 
%\end{figure}
The experimental results are reported in Table~\ref{AblationStudy}. 
Taking VGG-16 for illustration, it can be observed that without using IRM, IMS has a 1.1\% degeneration.  
By comparing the performance between ``w/o IB" and ours, we observe that the IB loss may impact the performance of our approach by 2.8\% in terms of accuracy. Similar results can be observed in other backbones. 
This indicates that the IMS can better guide the learning process to find the invariant representation and avoid the interference of redundant features.   

\begin{table}[hbpt]\small
\centering
\caption{Ablation studies on CDLT-cd dataset. Different backbones are used to evaluate the generalization of the IMS.}
%\vspace{-0.4cm} 
  \label{AblationStudy}
 \begin{tabular}{c|c|  ccc}  
 \bottomrule 
  	 Backbone    & Baseline  & w/o IB &  w/o IRM & Ours\\  
    \hline
     VGG-16      & 53.48  & 55.23 & 56.91 & \bf 58.04  \\
     ResNet-50   & 68.05  & 68.91 & 69.04 & \bf 69.63  \\      
     ViT-B-16     & 68.47  & 69.42   & 70.01   & \bf 70.80 \\
  \hline  
\end{tabular}
\vspace{-0.5cm} 
\end{table}

\begin{figure*}[htbp]
  \begin{center}
  \includegraphics[width=0.99\textwidth]{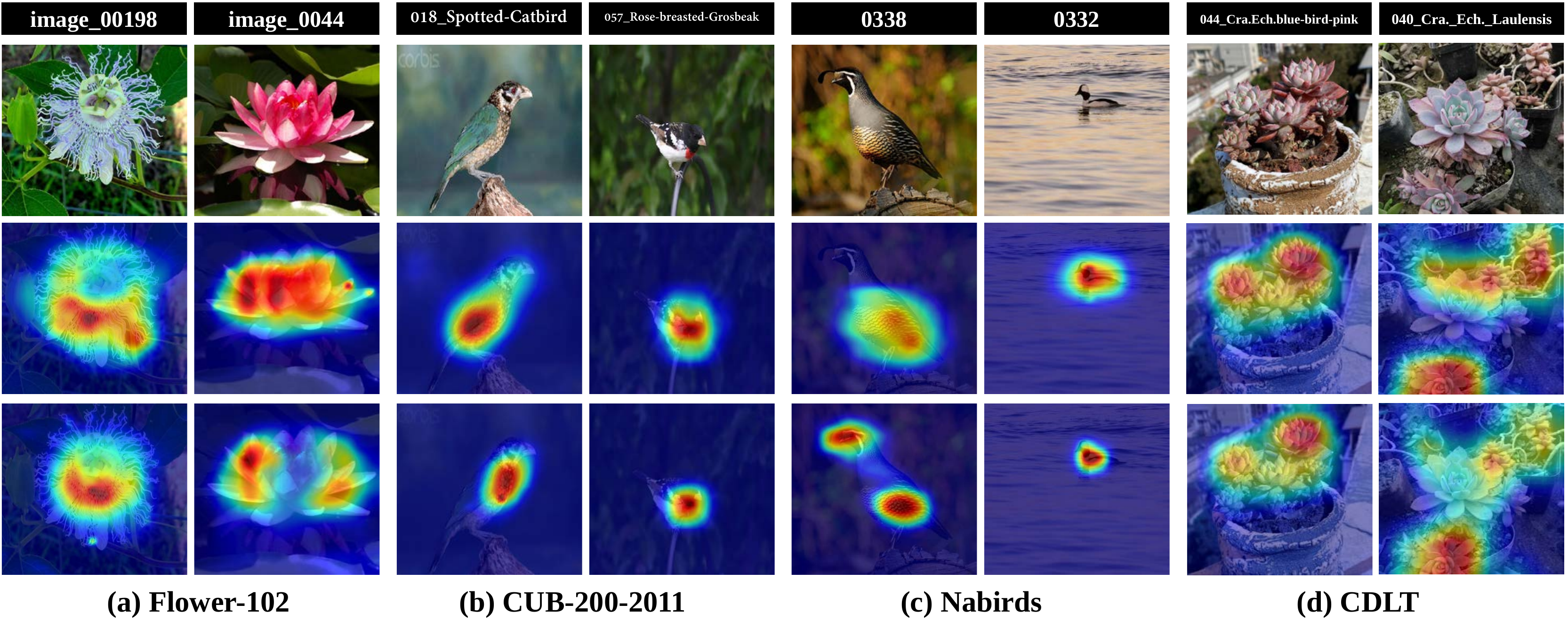}
  \end{center}
	\setlength{\abovecaptionskip}{-0.1cm} 
\vspace{-0.4cm} 
  \caption{Grad-CAM~\cite{selvaraju2017grad} visualization. The first row is the original image, the second and third rows are the attention visualization of baseline and IMS, respectively. Our method reliably identifies instance regions and extracts the minimum but sufficient features for learning, even in challenging background contexts.}
  \label{AB}
\vspace{-0.1cm} 
\end{figure*}

We have demonstrated the effectiveness of our IMS by visualizing our localization in Fig.~\ref{AB}. The first row shows the original image in a natural environment, the second row uses ResNet-50 as the baseline, and the last row displays our result.
In Fig.~\ref{AB}(a), both the baseline and our method have identified the main objects of the flowers. However, it is evident that our decision-making process reduces the use of redundant features, especially in the second instance where the flower contains a lot of duplicated features (e.g., red petals), and our method only utilizes a part of them to make the correct discrimination. 
Fig.~\ref{AB}(b) shows two bird images. While the baseline model correctly identified the instance, it used the entire visible area of the bird’s belly or back to make its decision. On the other hand, our method successfully classified the instance using only a minimal portion of the bird’s belly or back. This improvement demonstrates that our model learned to identify the most essential features of the instance, resulting in a more efficient representation. 
In Fig.~\ref{AB}(c), for the first image, the baseline easily found the instance and correctly identified its subclass because the instance occupied most of the image and is in a relatively ideal display state. However, our method discovered more discriminative features, which was beneficial as it enables our algorithm to handle complex scenes (e.g., partial occlusion) more confidently. In the second image, compared to the baseline, our method was still able to find the instance more accurately despite it only occupying a small part of the image. 
In Fig.~\ref{AB}(d), the first image contains several identical instances. Comparing the baseline with our method, we can see that IMS uses fewer regions to make the correct decision. In the second image, there are background instances with colors similar to the target, but they do not belong to the same subclass. By examining the heatmap, it is clear that the baseline mistakenly identified the instances in the background as the target to be classified. In contrast, IMS significantly improves this issue by better learning the invariant features of the instance. The visualization results indicate that our IMS exhibits good consistency and can effectively learn the minimal but sufficient and invariant features of instances from different environments.

To further illustrate the effect of our method, we quantitatively analyze the mutual information between each layer of features and input image $\mathbf{x}$ and label $y$ in the model. The result is shown in Fig.~\ref{IXTY}, where the blue line and the orange line represent the results of the original model~(with ERM) and IMS respectively. As can be seen, our method compresses the information of $\mathbf{x}$ and $\Phi(\mathbf{x})$, while the information of $\Phi(\mathbf{x})$ and $y$ remains the same. This suggests that our method does indeed reduce the expression of redundant features during training.

\begin{figure}[htbp]
  \begin{center}
  \includegraphics[width=0.97\textwidth]{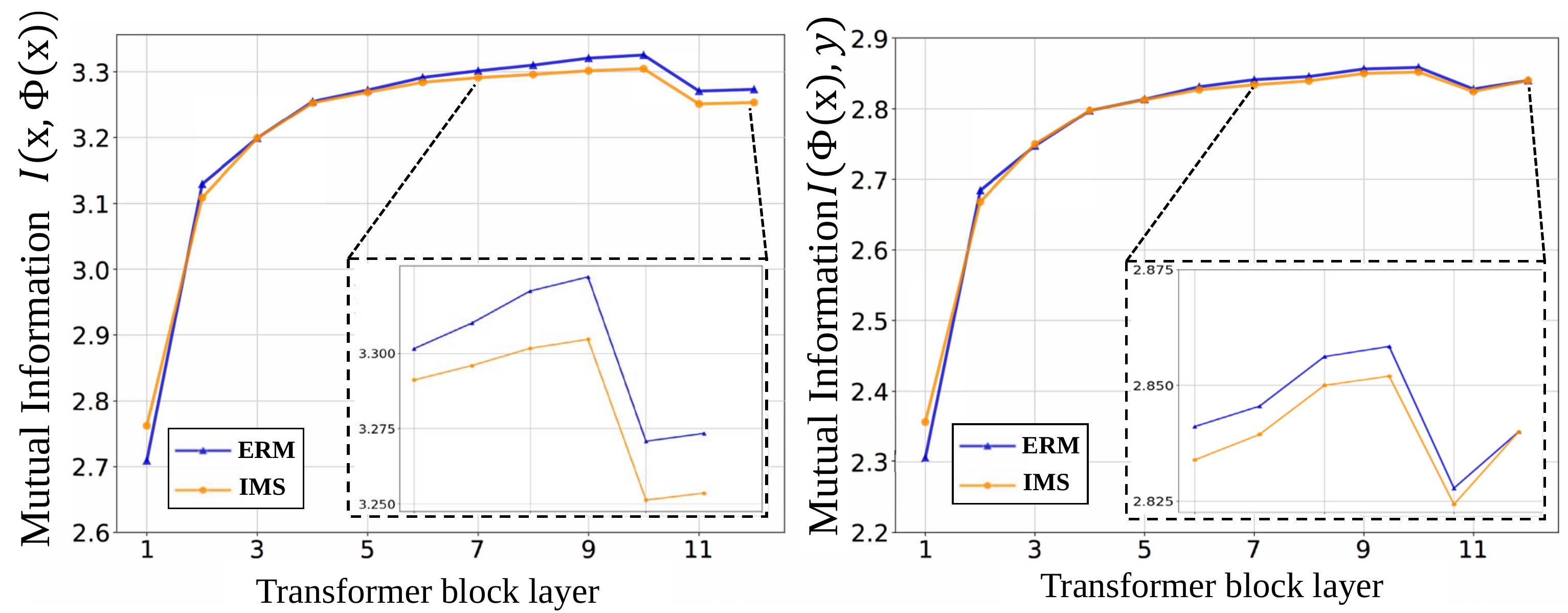}
  \end{center}
	\setlength{\abovecaptionskip}{-0.1cm} 
\vspace{-0.4cm}
\caption{Mutual information values of $I(\mathbf{x};\Phi(\mathbf{x})$ (left) and $I(y;\Phi(\mathbf{x})$ (right) in different layers. IMS compresses $I(\mathbf{x};\Phi(\mathbf{x})$ and remains the same amount of $I(y;\Phi(\mathbf{x})$ with respect to ERM.}
%The left and right graphs show the result of $I(\mathbf{x},\mathbf{t})$ and $I(\mathbf{t}, y)$, respectively. $\mathbf t$ is the feature representation without redundant features.
%The left and right graphs show the result of class token with input $\mathbf{x}$, and the class token with label $y$, respectively. }
\label{IXTY}
\vspace{-0.2cm} 
\end{figure}

\section{Conclusion}
Identifying the subtle, invariant, and most succinct representations is the key towards reliable fined-grained visual categorization (FGVC). 
In this paper, we propose an information-theoretic solution that combines the general ideas of invariant risk minimization (IRM) and information bottleneck (IB) principle to achieve this goal, by simply introducing two extra regularization terms (into existing learning objectives) without any assumptions on data/feature distributions or modification of network architectures.
Experimental results validated the prevalence of distributional shift in FGVC data and the non-Gaussian nature of deep neural network features. With a proper choice of hyper-parameters (e.g., the coefficients of two regularization terms), our IMS can always achieve state-of-the-art (SOTA) performance across five FGVC datasets with different backbones. We also validated, qualitatively and quantitatively, that our IRM is more likely to concentrate on invariant features and compress redundancy.

\section{Acknowledge}
This work was supported in part by the National Key R\&D Program of China 2022YFC3301000, in part by the Fundamental Research Funds for the Central Universities, HUST: 2023JYCXJJ031.
\begin{appendices}

\section{Further Analysis of Data Distribution}
To demonstrate the objective existence of distributional shift, we also conducted experiments using pre-trained ViT to replicate the experiments shown in Figure~\ref{Shift_quantization}. The experimental results are shown in Figure~\ref{DifferentMMD_Vit}.

\begin{figure}[htbp]
\setcounter{figure}{0}
\renewcommand{\thefigure}{A.\arabic{figure}}
  \begin{center}
  \includegraphics[width=0.97\textwidth]{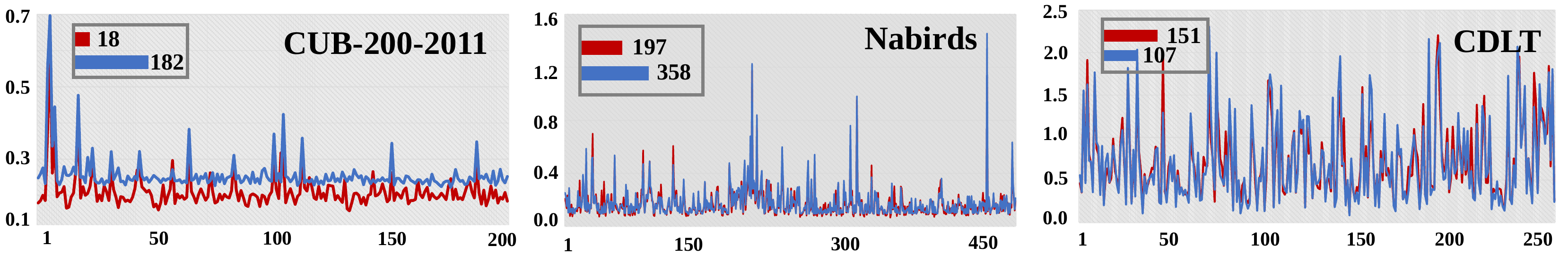}
  \end{center}
	\setlength{\abovecaptionskip}{-0.1cm} 
\vspace{-0.4cm}
\caption{Quantitative analysis of feature distribution under the ViT backbone network.}
\label{DifferentMMD_Vit}
\vspace{-0.2cm} 
\end{figure}
It can be observed that compared to the results of ResNet-50, there is a slight reduction in the occurrence of distributional shift. This is attributed to the self-attention mechanism of ViT, which makes the features captured by the model more robust, as evidenced by higher accuracy on different datasets. However, despite these improvements, distributional shift remains a commonly encountered issue, particularly evident in the Nabirds and CDLT datasets.

Since the requirement of partitioning the training dataset into different domains for the proposed method in this paper, we also employ MMD to measure the distance between different domains in order to assess the effectiveness of the IRM learning conditions.
In this process, we use the partition results of two domains for illustration, and the experiments also employ ViT for feature extraction, implemented using basic k-means.
The results are reported in Figure~\ref{MMDinDomain_Vit}.

\begin{figure}[htbp]
\setcounter{figure}{1}
\renewcommand{\thefigure}{A.\arabic{figure}}
  \begin{center}
  \includegraphics[width=0.97\textwidth]{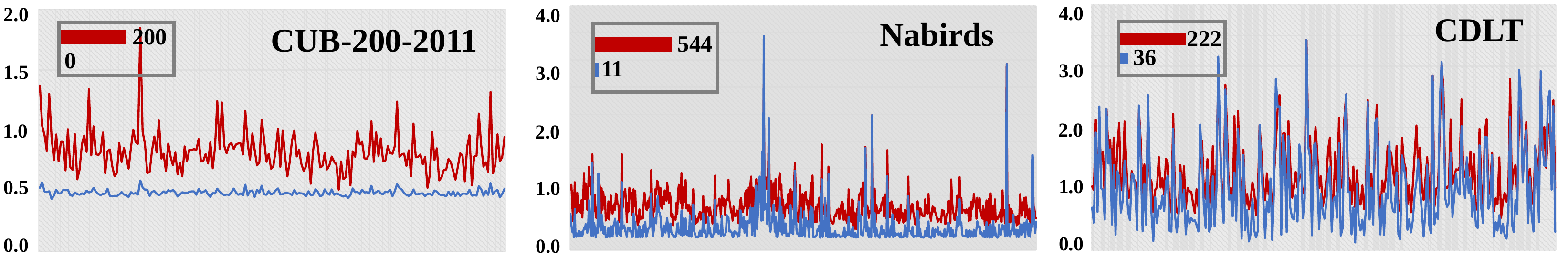}
  \end{center}
	\setlength{\abovecaptionskip}{-0.1cm} 
\vspace{-0.4cm}
\caption{Quantitative analysis of data distribution in different domains, based on ViT features.}
\label{MMDinDomain_Vit}
\vspace{-0.2cm} 
\end{figure}
It can be observed that our partitioning strategy is effective. On the CUB dataset, the training set is divided into two domains with no overlap. Even for the complex distribution of CDLT, the predefined objective is well achieved.

\section{Comparison with Decorr Method}
We compared our approach with the Decorr method~\cite{liao2022decorr}, which is currently considered the SOTA environment partitioning approach. The results of the comparison are presented in Table~\ref{Comparison_Decorr}.

\begin{table}[hbtp]\footnotesize
\setcounter{table}{0}
\renewcommand{\thetable}{B.\arabic{table}}
\setlength{\abovecaptionskip}{-0.04cm}
 \setlength{\tabcolsep}{7pt}
 \begin{center}
\caption{Comparison with Decorr. The batchsize~(Bs) is set to 24.}
\label{Comparison_Decorr}
 \begin{tabular}{c|ccccc}  
\hline
  $ $    & $K$=2 & $K$=3 & $K$=4 & $K$=5 & $K$=6    \\
   \hline
  Decorr & 90.84 & 90.62 & 90.54 & 90.41 & 90.20  \\
  Ours   & 90.89 & 90.73 & 90.84 & \bf 91.13 & 90.89  \\
  \hline  
\end{tabular} \end{center}
\vspace{-0.4cm}  
\end{table}

As can be seen, when the number of environments is small, our results are very close. When the number of environments is large, Doccer does not show any performance improvement or even performs worse. One possible reason for this is the rigid partitioning method employed by Doccer imposes limitations on the number of images per domain. When the number of images is small, it becomes challenging for IRM to learn the invariant features effectively.

\section{Examples From CDLT}
CDLT is a self-constructed dataset, and we demonstrate a part of the data in Figure~\ref{AppendixCDLT}. Each image belongs to a different subclass, and it can be observed that certain subclasses exhibit similar visual features, which aligns with the core challenge of fine-grained tasks. The complete dataset will be released as an open-source resource in the near future.
\begin{figure}[htbp]
\setcounter{figure}{0}
\renewcommand{\thefigure}{C.\arabic{figure}}
  \begin{center}
  \includegraphics[width=0.97\textwidth]{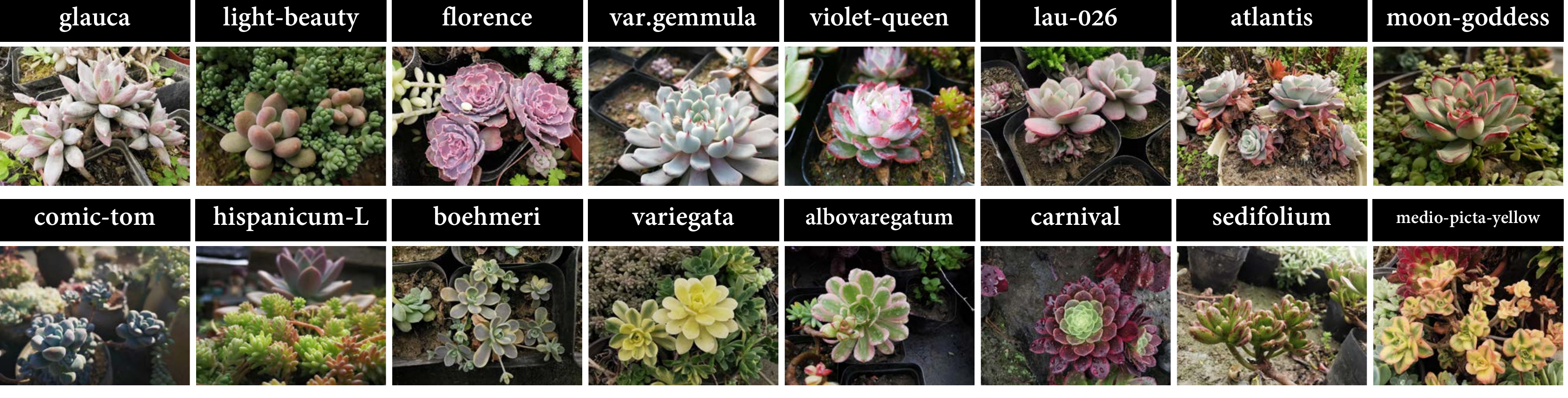}
  \end{center}
	\setlength{\abovecaptionskip}{-0.1cm} 
\vspace{-0.4cm}
\caption{Examples from CDLT.}
\label{AppendixCDLT}
\vspace{-0.2cm} 
\end{figure}

\section{Experiments on Painted Forms of Data}
We also investigated the generalization capability of IMS across different forms of data. 
Specifically, we trained the model using natural scene bird images (CUB-200-2011) and conducted testing using the painted forms of birds (CUB-200-Paintings)~\cite{wang2020progressive}. 
The results of the comparison are presented in Table~\ref{PaintedForms}.

\begin{table}[hbtp]\footnotesize
\setcounter{table}{0}
\renewcommand{\thetable}{D.\arabic{table}}
\setlength{\abovecaptionskip}{-0.04cm}
 \setlength{\tabcolsep}{7pt}
 \begin{center}
\caption{Experiment results on painted forms of data.}
\label{PaintedForms}
 \begin{tabular}{c|ccccc}  
\hline
  Method  & ResNet-50 &ViT-B-16 & PAN~\cite{wang2020progressive} & LADS~\cite{dunlap2022using}  & IMS(ViT)      \\
   \hline
   Acc.   & 47.88  & 66.78  & 67.40 & 66.18 & \bf 71.25  \\
  
  \hline  
\end{tabular} \end{center}
\vspace{-0.4cm}  
\end{table}   

It can be observed that when there is a more pronounced distributional difference between the training and test datasets, classical backbone networks like ResNet-50 and ViT-B-16 experience significant performance degradation. 
A comparison with Table~\ref{comparison_AU} reveals that the losses decreased by 37.62\% and 23.62\% respectively, revealing the limitations of existing backbone models in addressing distributional shifts.
PAN~\cite{wang2020progressive} achieves performance enhancement by hierarchically aligning the distribution between subcategories across Species, Genera, Families, and Orders levels using source domain instance labels.
LADS~\cite{dunlap2022using} attempted to introduce the text modality, leveraging the knowledge of pre-trained large models. 
In comparison to state-of-the-art methods, our approach still achieves a performance gain of 3.75\% by solely utilizing label information.
This reveals the effectiveness of our IMS in compressing redundant features and achieving the learning of invariant features across various data forms. These findings provide valuable insights into enhancing the model's robustness and applicability in diverse scenarios.

\end{appendices}

 \bibliographystyle{elsarticle-num} 
 \bibliography{CVIU_ref}

% \end{thebibliography}
\end{document}